\documentclass[conference]{IEEEtran}
\IEEEoverridecommandlockouts

\usepackage{caption} 
\usepackage{amsmath}
\usepackage{booktabs}       
\usepackage{multirow}
\usepackage{hyperref}
\usepackage{balance}
\usepackage{graphicx}
\usepackage[style=ieee,mincitenames=1,maxcitenames=2,maxbibnames=10,doi=false,isbn=false,url=false,eprint=false]{biblatex}
\DeclareSourcemap{
  \maps[datatype=bibtex, overwrite]{
    \map{
      \step[fieldset=address, null]
      \step[fieldset=location, null]
    }
  }
}

\begin{document}
\title{Advancements in Translation Accuracy for Stereo Visual-Inertial Initialization}

\author{\IEEEauthorblockN{1\textsuperscript{st} Han Song}
\IEEEauthorblockA{\textit{Computer Science} \\
\textit{University of Southern California}\\
California, USA \\
hsong427@usc.edu}
~\\
\and
\IEEEauthorblockN{2\textsuperscript{nd} Zhongche Qu*}
\IEEEauthorblockA{\textit{Computer Science} \\
\textit{Columbia University}\\
New York, USA \\
zq2172@columbia.edu}

~\\
\and
\IEEEauthorblockN{3\textsuperscript{rd} Zhi Zhang}
\IEEEauthorblockA{\textit{Computer Science} \\
\textit{New York University}\\
New York, USA \\
zz2310@nyu.edu}
~\\

\and
\IEEEauthorblockN{4\textsuperscript{th} Zihan Ye}
\IEEEauthorblockA{\textit{Technopreneurship and Innovation} \\
\textit{Nanyang Technological University}\\
Singapore, Singapore \\
yezi0004@e.ntu.edu.sg}

\and
\IEEEauthorblockN{5\textsuperscript{th} Cong Liu}
\IEEEauthorblockA{\textit{Computer Science} \\
\textit{Harbin Institute of Technology}\\
Shenzhen, China \\
liucong@stu.hit.edu.cn}
}


\maketitle

\begin{abstract}
As the current initialization method in the state-of-the-art Stereo Visual-Inertial SLAM framework, ORB-SLAM3 has limitations. Its success depends on the performance of the pure stereo SLAM system and is based on the underlying assumption that pure visual SLAM can accurately estimate the camera trajectory, which is essential for inertial parameter estimation. Meanwhile, the further improved initialization method for ORB-SLAM3, known as Stereo-NEC, is time-consuming due to applying keypoint tracking to estimate gyroscope bias with normal epipolar constraints. To address the limitations of previous methods, this paper proposes a method aimed at enhancing translation accuracy during the initialization stage. The fundamental concept of our method is to improve the translation estimate with a 3 Degree-of-Freedom (DoF) Bundle Adjustment (BA), independently, while the rotation estimate is fixed, instead of using ORB-SLAM3's 6-DoF BA. Additionally, the rotation estimate will be updated by considering IMU measurements and gyroscope bias, unlike ORB-SLAM3's rotation, which is directly obtained from stereo visual odometry and may yield inferior results when operating in challenging scenarios.
We also conduct extensive evaluations on the public benchmark, the EuRoC dataset, demonstrating that our method excels in accuracy.
\end{abstract}


\section{INTRODUCTION}
The combination of cameras and Inertial Measurement Units (IMUs) is highly effective for environmental sensing. Cameras capture a rich, detailed representation of the environment, while IMUs measure acceleration and angular velocity, providing robustness against fast motion and environments with little texture, as well as resilience to variations such as changes in lighting or motion blur. Together, these sensors complement each other perfectly. By harnessing both cameras and IMUs, Visual-Inertial SLAM (VI-SLAM) systems can accurately determine the metric six degrees-of-freedom (DOF) state~\cite{10503743,9197280, yang2023multivisualinertial}. Meanwhile, VI-SLAM systems are widely used in various fields and applications where accurate localization and mapping are crucial. These include autonomous vehicles~\cite{9340939, mescheder2019occupancy, 9788114, 9749965}, autonomous robots~\cite{feng2023robotic,hoxha2023robotic, 7759291, 10048510}, augmented reality (AR) and virtual reality (VR)~\cite{VINS, ORB2,Ye_2023_CVPR}, manipulation~\cite{gao2023autonomous} and underwater applications~\cite{xanthidis2022towards} such as marine archaeology, computer vision~\cite{shi2022pairwise}. Additionally, VI-SLAM systems are essential for effective path planning~\cite{9387081,10048510}, 3D reconstruction~\cite{xanthidis2021towards, 10160266, feng2023subsurface, Chen_2023_CVPR,Cai_2023_CVPR,Batsos_2018_CVPR,9995035} and plane reconstruction~\cite{chen2023planarnerf, Liu_2022_CVPR}.

However, while monocular VI-SLAM systems, which utilize a single camera and IMU, are cost-effective, they encounter several limitations, such as up-to-scale maps and a restricted field of view (FOV). Consequently, roboticists are increasingly adopting stereo VI-SLAM systems that use a stereo camera setup. This configuration offers enhanced capabilities, enabling the reconstruction of 3D geometry even without requiring camera motion.

There are many factors that impact the performance of stereo VI-SLAM systems in terms of accuracy. Initialization is a critical issue that affects the accuracy and usability of VI-SLAM. Moreover, initialization is the first step in running VI-SLAM because it is necessary to recover gravity, initial velocity, acceleration, and gyroscope biases, all of which are essential for the effective use of stereo VI-SLAM.  A good initialization is crucial for stereo VI-SLAM, requiring not only the accurate recovery of scale, gravity, initial velocity, acceleration, and gyroscope biases to enable the use of VI-SLAM, but also providing a solid initial value that serves as a good seed for the stereo VI-SLAM system.

Similar to the initialization methods used in monocular VI-SLAM systems, as described in various studies~\cite{Kneip, Martinelli, Kaiser, DongSi, jointviinit, viorb, weibovio, VINS,inertialonlyinit,EDI}, stereo VI-SLAM initialization can be divided into two categories: joint approaches~\cite{DongSi, openvins}  and disjoint approaches~\cite{weibo_stereo, ESVIO, qin2019general, ORBSLAM3TRO, wang2024stereo}.
The key difference between joint and disjoint methods is whether an additional Structure-from-Motion (SfM) problem is required, with inertial parameters subsequently derived based on trajectory from pure visual SLAM. 
In this paper, we focus on the disjoint method because it typically yields more accurate results than the joint method. The joint method often overlooks the gyroscope bias in its closed-form solution, which can lead to limited accuracy and is computationally expensive~\cite{EDI}.

As an initialization method for stereo VI-SLAM, VINS-Fusion~\cite{qin2019general}, building on VINS-Mono~\cite{VINS}, adopts a slightly different approach by jointly estimating velocity, gravity vector, and scale via visual-inertial bundle adjustment, rather than addressing them separately. In a similar vein, ORB-SLAM3~\cite{ORBSLAM3TRO} integrates the inertial-only optimization-based approach ~\cite{inertialonlyinit}into their stereo VI-SLAM system. Meanwhile, Stereo-NEC~\cite{wang2024stereo} enhances ORB-SLAM3 by improving both rotation and trajectory estimation. This is achieved by utilizing stereo normal epipolar constraints to determine the initial gyroscope bias and employing it to initiate a Maximum A Posteriori (MAP) problem for further refinement of inertial parameters. Additionally, it separates the estimation of rotation using IMU integration, leveraging precise rotation estimates to improve translation estimation through 3-DoF bundle adjustment.

While ORB-SLAM3 generally assumes accurate camera trajectory estimation in scenarios with sufficient baseline between consecutive frames~\cite{wang2024stereo}, its initialization also suffer in challenging conditions, such as during intense or pure rotation, leading to decreased accuracy and robustness. Additionally, while Stereo-NEC improves upon the methods used in ORB-SLAM3, it suffers from significant runtime delays due to the feature matching required when estimating gyroscope bias through eigenvalue-based optimization.

This need highlights the requirement for methods that can deliver improved performance in term of accuracy while mataining a comparable run-time speed. In response, we propose a method called ETA, which focuses on enhancing translation accuracy during the initialization stage. The core principle of our method is to refine the translation estimate using a 3 Degree-of-Freedom (DoF) Bundle Adjustment (BA), conducted independently, while keeping the rotation estimate fixed, as opposed to employing ORB-SLAM3's 6-DoF BA. Furthermore, unlike ORB-SLAM3, where the rotation is derived directly from stereo visual odometry and might perform suboptimally in tough scenarios, our approach updates the rotation estimate by taking into account IMU measurements and gyroscope bias. Additionally, our method aims to achieve performance comparable to Stereo-NEC while maintaining a runtime similar to that of ORB-SLAM3.

\section{PRELIMINARIES}
\subsection{Inertial Residual}
The inertial residual, \( r\mathcal{L}_{\text{k-1},\text{k}} \), at time \text{k} is defined as:
\begin{equation*}\small
\begin{aligned}
r\mathcal{L}_{\text{k-1},\text{k}} &= \left [r\Delta \textbf{R}_{\text{k-1},\text{k}}, r\Delta \textbf{V}_{\text{k-1},\text{k}}, r\Delta \textbf{t}_{\text{k-1},\text{k}} \right]^\top\\
r\Delta \textbf{R}_{\text{k-1},\text{k}} & = \log (\Delta \textbf{R}_{\text{k-1},\text{k}}^\top\textbf{R}_{\text{k-1}}^\top\textbf{R}_{\text{k}}) \\
r\Delta \textbf{V}_{k-1,k} &= \textbf{R}_{\text{k-1}}^\top(\textbf{V}_\text{k} - \textbf{V}_\text{k-1} - \textbf{g}\Delta t_{\text{k-1},\text{k}} - \Delta\textbf{V}_{\text{k-1},\text{k}}) \\
r\Delta \textbf{t}_{k-1,k} &=  \textbf{R}_{\text{k-1}}^\top(\textbf{t}_{\text{t}} - \textbf{t}_{\text{t-1}} - \textbf{V}_{text{k}}\Delta t_{\text{k-1},\text{k}} - \frac{1}{2}\textbf{g}\Delta t_{\text{k-1},\text{k}}^2) - \Delta \textbf{t}_{k-1,k}
\end{aligned}
\end{equation*}
where \( \textbf{R}_\text{k-1} \) and \( \textbf{R}_\text{k} \) denote the rotations relative to the world frame, and \( \Delta \textbf{R}_{\text{k-1},\text{k}}, \Delta \textbf{V}_{\text{k-1},\text{k}}, \) and \( \Delta 
\textbf{t}_{\text{k-1},\text{k}}\) are the IMU preintegration of rotation, velocity, and position measurements receptively and \textbf{g} is a gravity vector represented in world frame. $\Delta 
\textbf{t}_{\text{k-1},\text{k}}$ represents an interval between two consecutive frames.
\subsection{Visual Residual}
The visual residual, \( r\mathcal{L}_{reproj} \), at time \text{k} is defined as:
\begin{equation*}
r\mathcal{L}_{reproj} = \textbf{x}^{i}- \pi_{(\cdot)}(\textbf{R}_{\text{c}_{\text{k}},\text{w}}\textbf{X}^{i}+\textbf{t}_{\text{c}_{\text{k}},\text{w}})
\end{equation*}
\begin{align*}\small
\pi_m\left(
\left[\begin{matrix} 
X \\
Y \\
Z
\end{matrix}\right]\right) = \left[\begin{matrix} 
f_x\frac{X}{Z} + c_x \\
f_y\frac{Y}{Z} + c_y
\end{matrix}\right],
\pi_s\left(
\left[\begin{matrix} 
X \\
Y \\
Z
\end{matrix}\right]\right) = \left[\begin{matrix} 
f_x\frac{X}{Z} + c_x \\
f_y\frac{Y}{Z} + c_y \\
f_x\frac{X-b}{Z} + c_x
\end{matrix}\right], 
\end{align*}
where $\textbf{X}^{i}$ represents a 3D point in the world frame, derived through stereo matching given a known baseline or triangulation, while $\textbf{x}^{i}$ denotes the corresponding 2D feature. The functions $\pi_{(\cdot)}$ refer to the reprojection processes, with $\pi_m$ for monocular and $\pi_s$ for rectified stereo reprojection. $[X, Y, Z]^\top$ represents a 3D point coordinate, $f_x$ and $f_y$ are focal length, $c_x$ and $c_y$ are the principal point and $b$ is the baseline of a stereo camera.
\subsection{Visual-Inertial Residual}
Considering the combination of visual and inertial residuals, we can formulate the objective function for the entire Visual-Inertial SLAM (VI-SLAM) system as follows:
\begin{equation*}
\begin{aligned}
\textbf{e} & = r\mathcal{L}_{
\text{k-1},\text{k}} + r\mathcal{L}_{reproj} \\
\min_{\mathcal{X}} &= \sum{||r\mathcal{L}_{\text{k-1},\text{k}}}||^2_{\Sigma_{\text{k-1},\text{k}}} + \sum{||r\mathcal{L}_{reproj}||^2_{\Sigma_{\text{k-1},\text{k}}}}
\end{aligned}
\end{equation*}
where \( \Sigma_{\text{k-1},\text{k}} \) represents the corresponding covariance matrix for the inertial residual, and \( \Sigma_{\text{k-1},\text{k}} \) for the visual residual. Additionally, \( \mathcal{X} \) denotes the state that the VI-SLAM system needs to estimate.
\section{Proposed Approach}\label{sec:vi_3dgs_method}
Our method is inspired by Stereo-NEC~\cite{wang2024stereo}, which demonstrate that translation accuracy with 3-DoF bundle adjustment, when combined with precise rotation, is more accurate than a 6-DoF bundle adjustment. Building on Stereo-NEC, we separately estimate rotation using IMU integration, leveraging precise rotation estimates to enhance translation estimation through 3-DoF bundle adjustment. 
The underlying concept of our method combines the ORB-SLAM3~\cite{ORBSLAM3TRO} method with the 3-DoF steps of Stereo-NEC to enhance initialization performance. We start by formulating a MAP problem for estimating inertial parameters and then proceed to update rotation estimation and then we use it to enhance translation estimation in term of accuracy.

\begin{figure}[!ht]
    \vspace{-0.1in}
    \centering
\includegraphics[width=0.85\columnwidth]{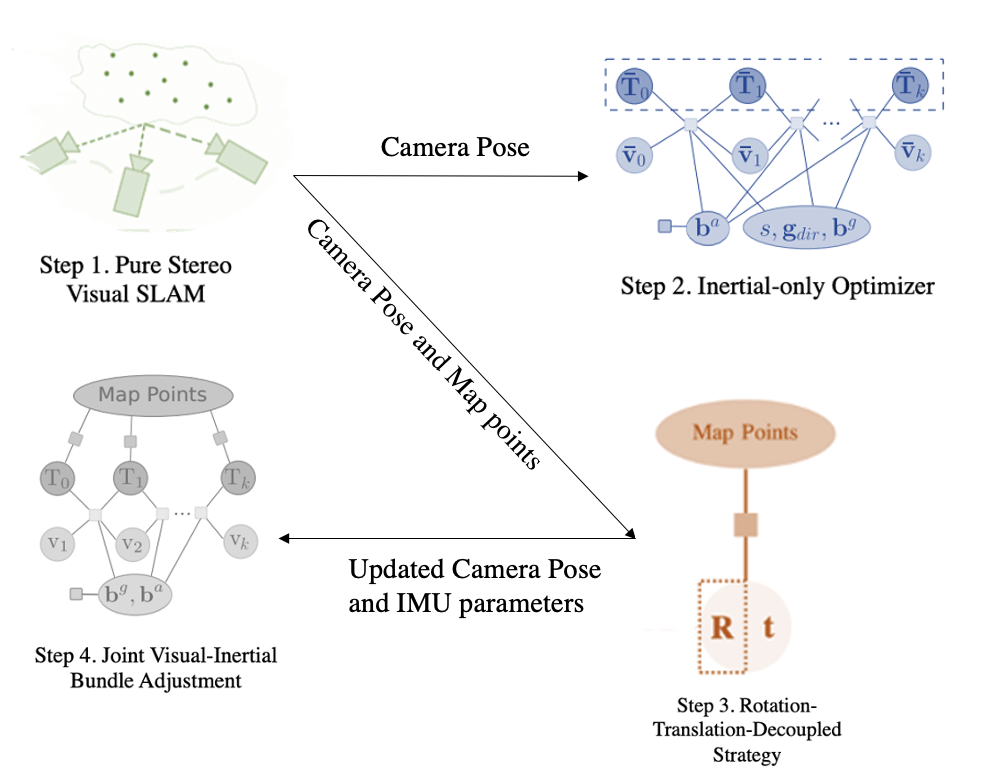}
    \caption{A diagram of the proposed pipeline. Initially, it runs pure stereo SLAM, followed by using an inertial-only optimizer to estimate inertial parameters. Next, it employs an efficient 3-DoF BA to enhance the translation estimates. Finally, it optimizes the entire set of visual and inertial parameters jointly through joint Visual-Inertial BA.}
    \label{fig:pipeline}
\end{figure}

Our method, as shown in ~\ref{fig:pipeline}, involves four steps aimed at deriving precise initial values for keyframes' poses and velocities, gravity direction, and IMU biases:
\begin{itemize}
    \item \textbf{Step 1. Pure Stereo Visual SLAM}: We recover the initial camera poses from stereo visual-only SLAM.
    \item \textbf{Step 2. Inertial-only Optimizer}: We estimate keyframes’ velocities, gravity direction, gyroscope bias and acceleration bias by solving  an inertial-only MAP estimation problem.
    \item \textbf{Step 3. Rotation-Translation-Decoupled Strategy}: We replease the camera rotation estimate by integrating gyroscope measurements with the gyroscope bias removed, and optimize the camera translation using 3-DoF BA from Stereo-NEC.
    \item \textbf{Step 4. Joint Visual-Inertial Bundle Adjustment}: 
  We utilize the solution derived from the previous steps as the initial estimate and construct a joint visual-inertial Maximum A Posteriori (MAP) problem to obtain optimal estimates for all inertial parameters, camera poses, and 3D landmark estimates.
\end{itemize}
\subsection{Pure Stereo Visual SLAM}\label{sec:step1}
Compared to monocular SLAM systems, Stereo SLAM systems provide a wider field of view (FOV) and offer the advantage of a known baseline. This allows Stereo SLAM systems to directly recover 3D landmarks with scale, even in static or pure rotation scenarios.

In this step, we run stereo SLAM with Bundle Adjustment using both Monocular and Stereo Constraints to estimate camera poses. The optimization employs the Levenberg-Marquardt method implemented in g2o, a non-linear optimization framework designed specifically for SLAM purposes.

Specifically, the Stereo SLAM system estimates camera poses, denoted as \([\textbf{R}_{\text{c}_{\text{k}},\text{w}},{\textbf{t}_{\text{c}_{\text{k}}, \text{w}}}]\), by minimizing the reprojection error between matched 3D points in the world frame and their corresponding keypoints, whether they are monocular or stereo keypoints, as shown is Equation~\ref{eq:translate_opt}:
\begin{equation}\small\label{eq:translate_opt}
\begin{aligned}
\{\textbf{R}_{\text{c}_{\text{k}},\text{w}},{\textbf{t}_{\text{c}_{\text{k}},\text{w}}}\} = \arg\min_{\textbf{R}_{\text{c}_{\text{k}},\text{w}}, \textbf{t}_{\text{c}_{\text{k}},\text{w}}}&\sum_{i\in\mathcal{M}}\rho\big(\Vert\textbf{x}^{i}- \pi_{(\cdot)}(\textbf{R}_{\text{c}_{\text{k}},\text{w}}\textbf{X}^{i}+\textbf{t}_{\text{c}_{\text{k}},\text{w}})\Vert^2_{\Sigma}\big) 
\end{aligned}
\end{equation}
where $\textbf{X}^{i}$ represents a 3D point in the world frame, derived through stereo matching given a known baseline, while $\textbf{x}^{i}$ denotes the corresponding 2D feature. The robust Huber cost function is denoted by $\rho$. The functions $\pi_{(\cdot)}$ refer to the reprojection processes, with $\pi_m$ for monocular and $\pi_s$ for rectified stereo reprojection. Additionally, $\Sigma$ signifies the covariance associated with the scale level of the keypoints within the pyramid, as detailed in~\cite{ORB2}.
\subsection{Inertial-only Optimizer}\label{sec:step2}
In this step, our goal is to derive estimates for keyframes' velocities, the direction of gravity, and IMU biases through Maximum A Posteriori(MAP), which incorporates an empirical prior residual. Practically, this is achieved by using ORB-SLAM3's Inertial-only MAP Estimation method. This method relies on keyframes within a sliding window and the inertial measurements recorded between these keyframes. 
\subsection{Rotation-Translation-Decoupled Strategy}\label{sec:step3}
Drawing inspiration from Stereo-NEC, we also apply a Rotation-Translation-Decoupled Strategy to enhance the accuracy of translation estimates. Stereo-NEC demonstrates that, after obtaining an optimal gyroscope bias from Step 2, updating each rotation estimate within the sliding window through IMU rotation integration, and optimizing each translation with only 3-DoF Bundle Adjustment (BA), can achieve more accurate results than the traditional 6-DoF BA. This traditional approach simultaneously considers rotation and translation estimates during the optimization phase.

Therefore, the first part of this step involves obtaining the rotation by using IMU integration, with the gyroscope bias removed following the Inertial-only optimizer step. Subsequently, we apply 3-DoF Bundle Adjustment (BA) to the translation, as shown in Equation~\ref{eq:translate_opt_2}:
\begin{equation}\small\label{eq:translate_opt_2}
\begin{aligned}
\textbf{R}_{\text{w},\text{c}_{\text{k}}} &=\textbf{R}_{\text{w},\text{b}_{\text{k}}}
\textbf{R}_{\text{b},\text{c}}\\
{\textbf{t}_{\text{c}_{\text{k}},\text{w}}}^* = \arg\min_{\textbf{t}_{\text{c}_{\text{k}},\text{w}}}&\sum_{i\in\mathcal{M}}\rho\big(\Vert\textbf{x}^{i}- \pi_{(\cdot)}(\textbf{R}_{\text{c}_{\text{k}},\text{w}}\textbf{X}^{i}+\textbf{t}_{\text{c}_{\text{k}},\text{w}})\Vert^2_{\Sigma}\big) 
\end{aligned}
\end{equation}
The rotation \( \textbf{R}_{\text{w},\text{b}_{\text{k}}} \) is obtained from IMU integration with the gyroscope bias removed, and \( \textbf{R}_{\text{b},\text{c}}\) represents the IMU-Camera extrinsic matrix derived from calibration and it is worth noting that \( \textbf{R}_{\text{w},\text{c}_{\text{k}}}\) is fixed during optimization.
\subsection{Joint Visual-Inertial Bundle Adjustment}\label{sec:step4}
Our final step involves applying Joint Visual-Inertial Bundle Adjustment, considering both inertial and visual residuals once we have a good estimation of inertial and visual parameters. This step aims to further refine the estimates of inertial parameters, camera poses, and 3D landmark positions through a joint visual-inertial optimization.
\section{EVALUATION}
\subsection{Experiment Results}
To assess accuracy across different sequences, we conducted an exhaustive initialization test. In this test, we initiated an initialization every 2.5 seconds for each sequence, resulting in the evaluation of 464 distinct initialization segments. Table~\ref{tab:table1} presents a comparison of rotation errors between our results and those obtained by ORB-SLAM3, while Table\ref{tab:table2} displays the absolute trajectory errors. Both tables include data with and without integration with visual-inertial bundle adjustment. Our method outperforms ORB-SLAM3 in several machine hall environments, specifically in the V1\_01 and V2\_03 machine halls in terms of rotation error, and in the V2\_02 machine hall in terms of absolute trajectory error. Overall, our method produces results that are comparable to those of ORB-SLAM3. Specifically, our results outperform those of ORB-SLAM3 on the Vicon Room 2-3(V2\_03\_difficult) dataset in both Absolute Trajectory Error and Rotation Error. This achievement is particularly noteworthy given the challenges presented by the Vicon Room 2-3(V2\_03\_difficult) dataset, which includes motion blur and illumination changes. These factors pose significant difficulties for the state estimator.
\begin{table}[!ht]
    \centering
    \vspace{0.20in}
    
    \begin{tabular}{@{}lcccc}
        \toprule  
        \multirow{3}{*}{\textbf{MHall \& VRoom}} &\multicolumn{2}{c}{\textbf{Rotation Error}} \\
 &\multicolumn{2}{c}{\textbf{Without VI-BA}} & \multicolumn{2}{c}{\textbf{With VI-BA}}  \\
& ORB-SLAM3 &Ours & ORB-SLAM3 &Ours \\
    \hline
     MH\_01\_easy &\textbf{0.083} &0.084 &0.020 &\textbf{0.019}\\
     MH\_02\_easy &\textbf{0.093} &\textbf{0.093} &\textbf{0.016} &\textbf{0.016}\\
     MH\_03\_medium &0.344 &\textbf{0.335} &0.036 &0.037\\
     MH\_04\_difficult &\textbf{0.165} &0.171 &\textbf{0.032} &0.035 \\
     MH\_05\_difficult &\textbf{0.153} &0.169 &\textbf{0.028} &0.030 \\
     V1\_01\_easy &\textbf{0.188} &0.193 &0.115 &\textbf{0.112}\\
     V1\_02\_medium &0.448 &\textbf{0.405} &0.065 &\textbf{0.064} \\
     V1\_03\_difficult &0.892 &0.924 &\textbf{0.095} &\textbf{0.095} \\
     V2\_01\_easy  &\textbf{0.177} &0.190 &0.042 &\textbf{0.040}\\
     V2\_02\_medium &\textbf{0.320} &0.328 &0.076 &\textbf{0.070} \\
     V2\_03\_difficult &0.987 &\textbf{0.910} &\textbf{0.214} &0.224 \\
    \hline
    Avg &0.350 &\textbf{0.345} &\textbf{0.067} &\textbf{0.067}\\
\bottomrule
    \end{tabular}
    
    \caption{Comparison of rotation error between ours results and ORB-SLAM3 results in without VI-BA and with VI-BA}
    \label{tab:table1}
\end{table}

\begin{table}[!ht]
    \centering
    \vspace{0.20in}
    \begin{tabular}{@{}lcccc}
        \toprule  
        \multirow{3}{*}{\textbf{MHall  VRoom}} &\multicolumn{2}{c}{\textbf{Absolute Trajectory Error}} \\
 &\multicolumn{2}{c}{\textbf{Without VI-BA}} & \multicolumn{2}{c}{\textbf{With VI-BA}}  \\
& ORB-SLAM3 &Ours & ORB-SLAM3 &Ours \\
    \hline
     MH\_01\_easy &\textbf{0.007} &\textbf{0.007} &\textbf{0.004} &\textbf{0.004}\\
     MH\_02\_easy &\textbf{0.006} &\textbf{0.006} &\textbf{0.004} &\textbf{0.004}\\
     MH\_03\_medium &\textbf{0.034} &\textbf{0.034} &\textbf{0.012} &\textbf{0.012}\\
     MH\_04\_difficult &\textbf{0.027} &0.028 &\textbf{0.014} &\textbf{0.014} \\
     MH\_05\_difficult &\textbf{0.023} &0.026 &\textbf{0.011} &0.012 \\
     V1\_01\_easy &\textbf{0.008} &\textbf{0.008} &0.006 &\textbf{0.005}\\
     V1\_02\_medium &0.022 &\textbf{0.020} &0.007 &\textbf{0.006}\\
     V1\_03\_difficult &\textbf{0.052} &0.054 &0.009 &\textbf{0.008} \\
     V2\_01\_easy  &\textbf{0.006} &\textbf{0.006} &\textbf{0.003} &\textbf{0.003}\\
     V2\_02\_medium &\textbf{0.019} &0.021 &0.007 &\textbf{0.006} \\
     V2\_03\_difficult &0.056 &\textbf{0.055} &\textbf{0.016} &\textbf{0.016} \\
    \hline
    Avg &\textbf{0.023} &0.024 &\textbf{0.008} &\textbf{0.008}\\
\bottomrule
    \end{tabular}
    
    \caption{Comparison of Absolute Trajectory Error error between ours results and ORB-SLAM3 results in without VI-BA and with VI-BA}
    \label{tab:table2}
\end{table}

\subsection{Environment setup and implementation}
The EuRoc dataset offers highly accurate rotational and transactional data for 11 Micro Air Vehicle(MAV) sequences, covering a range of flight conditions. This dataset includes synchronized visual inertial sensor units equipped with global shutter cameras and a MEMS inertial Measurement Unit(IMU) that provides angular rage and acceleration data. Additionally, it provides camera intrinsic parameters and camera-IMU extrinsic parameters. All experiments were conducted on an Intel i7-9700k desktop computer with 64 GB of RAM. To ensure an equitable  comparison between the initialization methods of ORB-SLAM3 and Stero-NEC, our method was integrated into ORB-SLAM3. In all our evaluations, the absolute trajectory error(ATE) was measured in meters without involving scale alignment. 
\begin{figure}[!ht]
    \vspace{-0.1in}
    \centering
\includegraphics[width=0.75\columnwidth]{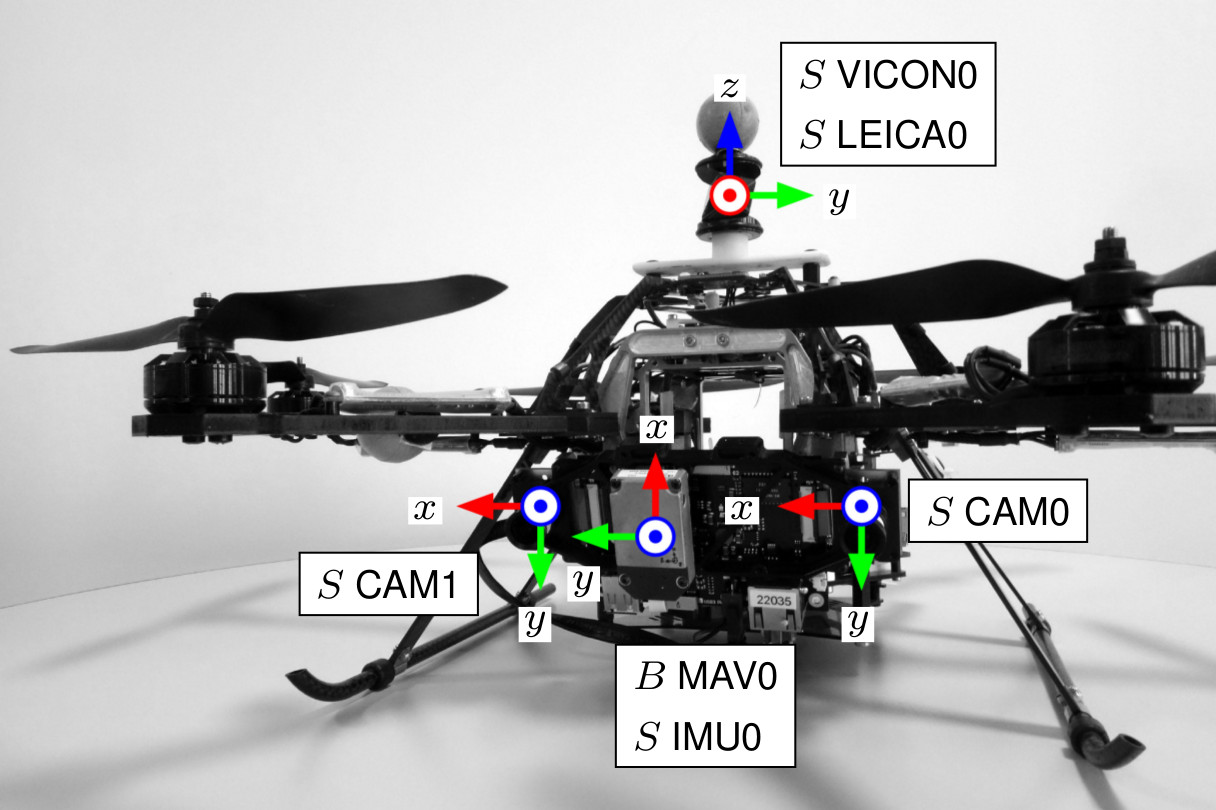}
    \caption{An Asctec Firefly hex-rotor aerial drone was used, equipped with a visual-inertial sensor unit comprising a camera and an IMU.}
    \label{fig:platform}
\end{figure}
\section{CONCLUSIONS}
In conclusion, this study addresses the limitations of the current initialization methods in the state-of-the-art Stereo Visual-Inertial SLAM framework, specifically ORB-SLAM3 and its improved version, Stereo-NEC. ORB-SLAM3's reliance on the pure stereo SLAM system for accurate camera trajectory estimation and inertial parameter estimation, along with Stereo-NEC's time-consuming keypoint tracking for gyroscope bias estimation, highlight the need for an enhanced approach. We propose a novel initialization method that improves translation accuracy by employing a 3 Degree-of-Freedom (DoF) Bundle Adjustment (BA) for the translation estimate while keeping the rotation estimate fixed. Unlike ORB-SLAM3’s 6-DoF BA, our method independently updates the rotation estimate by incorporating IMU measurements and gyroscope bias, thus addressing the deficiencies of direct rotation estimation from stereo visual odometry in challenging scenarios. Extensive evaluations on the EuRoC dataset demonstrate the superior accuracy of our method compared to ORB-SLAM3. Notably, our method outperforms ORB-SLAM3 on the V2\_03\_difficult dataset, which is particularly challenging due to factors such as motion blur and illumination changes that complicate state estimation. These results highlight the robustness and reliability of our approach in diverse and demanding environments. Our method not only enhances initialization performance but also shows significant potential for practical applications in real-world visual-inertial state estimation tasks.environments.
\printbibliography
\end{document}